\documentclass{article}

\usepackage{arxiv}

\usepackage[utf8]{inputenc} 
\usepackage[T1]{fontenc}    
\usepackage{hyperref}       
\usepackage{url}            
\usepackage{booktabs}       
\usepackage{amsfonts}       
\usepackage{nicefrac}       
\usepackage{microtype}      
\usepackage{lipsum}		
\usepackage{graphicx}
\usepackage[numbers]{natbib}
\usepackage{doi}
\newcommand{\degree}{$^{\circ}$}
\usepackage{makecell}
\usepackage{multicol}
\usepackage{multirow}
\usepackage{subcaption}

\title{Privacy Enhancement for Gaze Data Using a Noise-Infused Autoencoder}

\author{ 
    \href{https://orcid.org/0000-0002-7656-2662}{\includegraphics[scale=0.06]{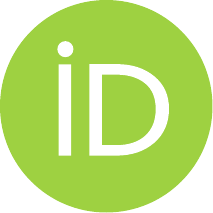}\hspace{1mm}Samantha Aziz} \\
	Department of Computer Science\\
	Texas State University\\
	San Marcos, Texas, USA\\
	\texttt{sda69@txstate.edu} \\
    \And	
    \href{https://orcid.org/0000-0001-7890-8842}{\includegraphics[scale=0.06]{orcid.pdf}
    \hspace{1mm}Oleg Komogortsev} \\
	Department of Computer Science\\
	Texas State University\\
	San Marcos, Texas, USA \\
	\texttt{ok@txstate.edu} \\
}

\renewcommand{\shorttitle}{Privacy Enhancement for Gaze Data Using a Noise-Infused Autoencoder}

\hypersetup{
pdftitle={Privacy Enhancement for Gaze Data Using a Noise-Infused Autoencoder},
pdfsubject={cs.HC},
pdfauthor={Samantha Aziz, Oleg Komogortsev},
pdfkeywords={eye tracking, privacy, deep learning, biometrics, user profiling, autoencoder},
}

\begin{document}
\maketitle

\begin{abstract}
 We present a privacy-enhancing mechanism for gaze signals using a latent-noise autoencoder that prevents users from being re-identified across play sessions without their consent, while retaining the usability of the data for benign tasks.
 We evaluate privacy-utility trade-offs across biometric identification and gaze prediction tasks, showing that our approach significantly reduces biometric identifiability with minimal utility degradation. 
 Unlike prior methods in this direction, our framework retains physiologically plausible gaze patterns suitable for downstream use, which produces favorable privacy-utility trade-off.
 This work advances privacy in gaze-based systems by providing a usable and effective mechanism for protecting sensitive gaze data.
 \end{abstract}


\section{Introduction}
Although its primary use has traditionally been a tool for academic research, eye tracking technology has recently emerged as a novel interaction modality on consumer devices across the market, including extended reality devices and smartphones.
On these platforms, eye tracking supports a range of applications that enhance a user's interaction experience, including gaze-based interaction~\cite{plopski2022eye}, improved social communication~\cite{courgeon2014joint}, and novel accessibility measures~\cite{hwang2014eye}.
This recent emergence onto the consumer market represents a substantial advancement of eye tracking's availability, where the widespread collection and utilization of gaze data allows more users to embrace the benefits of gaze-based applications.

However, the widespread adoption of eye tracking also raises significant privacy concerns, as gaze data also conveys sensitive information that can be exploited to infer personal characteristics about a user.
Eye tracking signals are produced by complex neurological processes that also dictate personal characteristics, such as a user's expression of their identity, gender, personality, and age~\cite{kroger}.
As a result, eye tracking signals are highly individualistic, and can be used to uniquely identify users.
Recent work in psychophysiological biometrics research demonstrates that users can be reliably uniquely identified among a large group of 9,000 users, and theorizes that users can be uniquely identified within a group of nearly 150,000 others~\cite{lohr2024establishing}.
Users who provide gaze data untrustworthy applications are inadvertently sharing uniquely identifying information, and it is possible for malicious parties who gain access to that data to reveal these personally identifying characteristics without the user's knowledge or consent. 

As gaze data becomes more commonly integrated within consumer-facing technologies, it is critical to develop privacy-preserving mechanisms that allow users to embrace the novel applications that eye tracking facilitates while preventing the unnecessary disclosure of personal traits that puts their privacy at risk. 
This work aims to enhance user privacy in gaze-based interaction systems, particularly in ubiquitous computing contexts where eye tracking data may be shared to third parties to enable vital applications.

Toward this end, we propose a method for privatizing gaze signals that is based on an autoencoder architecture with latent-space noise injection, which offers a practical method for enhancing user privacy in these settings by allowing for scalable privacy enhancement without the need for post-hoc filtering or unrealistic data abstractions.
Moreover, we demonstrate that this adaptation is particularly well-suited for preserving the usability of gaze signals in downstream tasks, such as gaze prediction, even under increasing levels of privacy.

The major contributions of this work are as follows:
\begin{enumerate}
\item The introduction and validation of a signal-level privacy-enhancing mechanism for gaze data using an autoencoder with latent noise injection that preserves key physiological characteristics of eye movements.
\item An empirical demonstration that our method provides a favorable privacy-utility trade-off, maintaining compatibility with real-world gaze-based tasks such as gaze prediction while substantially reducing the risk of biometric identification for users.
\end{enumerate}

\section{Background and Related Work}
\subsection{Privacy for Gaze Data}
Several approaches to enhancing privacy for gaze data have been proposed in literature, each targeting privacy enhancement for different representations of a user's gaze.
For example, some approaches transform gaze signals into another application-specific representation before applying a privacy-enhancing mechanism, such as scanpath images~\cite{liu, Fuhl2021} and sets of aggregated eye tracking features~\cite{steil}.
The most relevant prior work for privacy enhancement of eye tracking data involve mechanisms that directly perturb gaze signals, which can be useful in practical applications that operate directly on gaze signals~\cite{davidjohn2021, Hu2022_otus, wilson2024privacypreserving}.

One key challenge in this direction is that injecting noise directly into gaze signals can significantly degrade the signal's utility~\cite{davidjohn2021}, producing high levels of utility degradation relative to the amount of privacy gained.
To address this, our proposed autoencoder provides privacy enhancement by adding noise to the latent representation of gaze that is produced during the model's encoding process.
In doing so, the mechanism produces more desirable privacy-utility trade-offs relative to mechanisms that add noise to the gaze signal, giving the intended effect of obfuscating privacy-sensitive features without distorting the signal to the degree that it cannot be used for an intended utility.

\subsection{Autoencoders for Privacy Enhancement}
Autoencoders consists of two components: an encoder and a decoder.
The encoder is a neural network that transforms an input into an abstract, low-dimensionality latent representation.
The decoder reconstructs the original input from that latent representation. 
The reconstructed signal is similar to the original input, but may be be absent of certain details that were lost during the encoding process. 
Because the encoder must compress the input into a low-dimensional space, the model learns to capture the features of the data that are most important for faithful reconstruction.

Autoencoders have been applied for privacy protection in several other domains, obscuring sensitive user traits from images~\cite{Azizian2024, Jamshidi2024_learning_structure, Mirjalili2017, Liu2023, Fuhl2021}, audio signals~\cite{Perero2022}, tabular data~\cite{Mandal2022}, and data that is used for recommender systems~\cite{Deng2025, Liu2019, Ren2019}.
These approaches are summarized in Table~\ref{tab:prior-work}.
We propose that an autoencoder can be used to enhance privacy by obfuscating sensitive information in a gaze signal.
Because of the bottleneck structure of the latent representation, the autoencoder can be trained to accurately reconstruct the basic structure of an average gaze signal while discarding subject-specific details that may otherwise reveal sensitive characteristics.

The use of an autoencoder for eye tracking privacy has been explored in prior work~\cite{Fuhl2021}, but only when operating on images of a scanpath, which is a low-dimensional representation of gaze that is not practical for use in real-world settings that use gaze signals.
By designing an autoencoder-based privacy mechanism that can be applied directly to gaze signals, this work solves an inherently more difficult problem and produces a privacy enhancement solution that is more practical for real-world use.

\begin{table*}[]
    \caption{Summary of the Prior Work Using Autoencoders for Privacy Enhancement}
    \centering
    \begin{tabular}{c|c|c|c|c}
         Work & Utility & Privacy Target & Data Type & Privacy Demonstration \\
         \hline
         \cite{Azizian2024} & Object detection & Faces, License Plates & Images & Empirical \\
         \cite{Deng2025} & Recommender systems & User identity & Embedding Matrices & Formal \\
         \cite{Jamshidi2024_learning_structure} & Facial Expression & Gender & Face Images & Empirical \\
         \cite{Liu2019} & Recommender systems & Differential Privacy & Embedding Matrices & Formal \\
         \cite{Mirjalili2017} & Biometric Verification & Gender & Face Images & Empirical \\
         \cite{Ren2019} & Recommender systems & Differential Privacy & Embedding Matrices & Formal \\
         \cite{Liu2023} & Facial feature recognition & Biometric Identity & Face Images & Empirical \\
         \cite{Perero2022} & Speech recognition & Identity, gender, accent & Audio & Empirical \\
         \cite{Mandal2022} & Income & Gender & Tabular Data & Empirical \\
         \cite{Fuhl2021} & Scanpath generation & User attributes & Scanpath images & Empirical \\
    \end{tabular}
    \label{tab:prior-work}
\end{table*}

\section{Methodology}
\subsection{Problem Setting}
This investigation considers the problem of private data release to untrusted third parties for individual users. 
Consider a scenario where a user-generated gaze signal is captured and processed on a consumer device, like a virtual reality headset.
The data is then streamed to a third party to enable vital functionality, such as animating the user's virtual avatar in a shared virtual environment.

The third party is an honest-but-curious entity that secondarily uses this data to gain insights about the user that it would not have access to otherwise.
Namely, this third party uses this gaze data to re-identify users across different play sessions, perhaps to track their activity over time or serve them advertisements.
This constitutes a violation of the user's privacy, as they did not consent this secondary utilization of their gaze data.
The goal, then, is to obfuscate this personally identifying information from the gaze signal before it is broadcast to third parties, thereby safeguarding the user's privacy while allowing the data to be used for a benign stated purpose.
The proposed autoencoder mechanism acts as an intermediate layer between the device and third party applications, and applies privacy-enhancing transformations to eliminate personally identifying information from the gaze signal before sharing the data with untrusted third parties.

\subsection{Data Set}
\label{sec:data}
We use data from the publicly available GazeBase data set~\cite{Griffith2021_gazebase} to train and evaluate the privacy mechanism described in this investigation.
GazeBase is a large-scale, longitudinal eye movement data set that contains high-quality eye movement signals captured by an EyeLink 1000 eye tracking device.
Monocular gaze data was collected from the left eye at a sampling rate of 1000 Hz while participants completed a battery of tasks that involved following a dot on the screen (FXS, HSS, RAN), watching a video (VD1, VD2), reading an excerpt of text (TEX), and completing a gaze interaction task (BLG).

For this investigation, we split GazeBase into subject-disjoint training, validation, and testing sets to prevent subject-specific characteristics from biasing our evaluation. 
We train the autoencoder and biometric model using data from subjects who are only in Round 1 (186 subjects).
Subjects who were present in Round 2 (but not later), were used as the validation set (31 subjects).
The remainder of subjects (i.e., those present in Round 3), were set as the testing subjects (105 subjects).
When training and validating the model, we used all of the available data from Rounds 1-9 from each subject, except for data from the BLG task.
When evaluating the model, we use the Round 1 RAN data from users in the testing set.
We use this training and evaluation set to train both the privacy-enhancing autoencoder (Section~\ref{sec:autoencoder}) and the biometric model used for assessment~\ref{sec:ekyt}.

\subsection{Autoencoder Model}
\label{sec:autoencoder}

\subsubsection{Data Pre-Processing}
A gaze signal is a collection of ordered gaze samples $X = \{{(t_0, x_0, y_0), (t_1, x_1, y_1), ..., (t_n, x_n, y_n)}\}$, where each sample contains the timestamp ($t$) and the horizontal and vertical gaze position ($x,y$) in degrees of visual angle (\degree).
Gaze data is first downsampled from a sampling frequency of 1000 Hz to 250 Hz, to make the data better emulate a sampling frequency available in consumer-grade devices, then split into non-overlapping sequences of 64 milliseconds.
To prevent the model from reconstructing a gaze signal from missing data, all sequences containing invalid data (i.e., NaNs), were excluded from the training set (NaNs are not excluded at evaluation).

\begin{figure}
    \centering
    \includegraphics[width=0.4\linewidth]{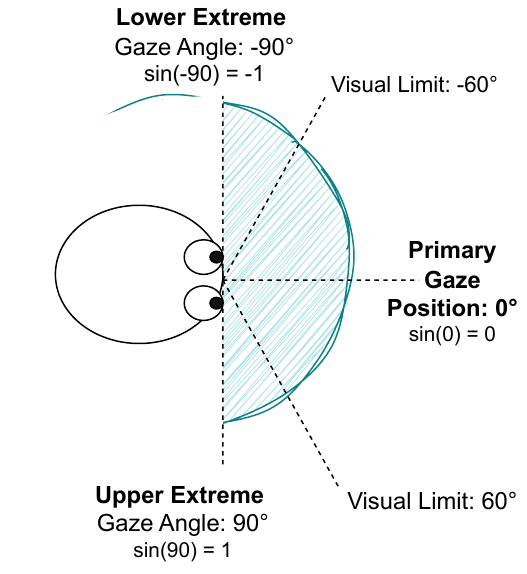}
    \caption{The intuition behind scaling gaze positions to a range of [-1,1] based on limitations of the human visual system.}
    \label{fig:data-preprocess}
\end{figure}

To encourage physiologically feasible outputs, we convert all gaze positions into their trigonometric values by applying the function $X' = (sin(x_t),sin(y_t))$ to each sample in the signal.
The intuition behind this, summarized in Figure~\ref{fig:data-preprocess}, involves the physical limitations of the human visual system that bound the maximum field of view to 180 degrees.
Assuming that the primary position of a user's gaze corresponds to 0\degree~horizontally, it is not physically possible for the gaze angle of an individual eye to exceed 90\degree~relative to the primary position.
Therefore, we can assume that all gaze data would only contain gaze samples in the range $\pm$90\degree~relative to the center of the user's head.
Because we expect all valid gaze positions to be within this range, it is straightforward to scale the horizontal and vertical components of the gaze signal with simple trigonometry.
In a practical sense, most gaze samples likely fall within the human visual limit of approximately $\pm$60\degree, but setting the absolute limit of gaze angles to $\pm$90\degree~allows us to more intuitively map gaze positions to trigonometric values.
This strategy also helps the model converge more smoothly and favors physiologically plausible outputs.

\subsubsection{Architecture}
\begin{figure*}
    \centering
    \includegraphics[width=\linewidth]{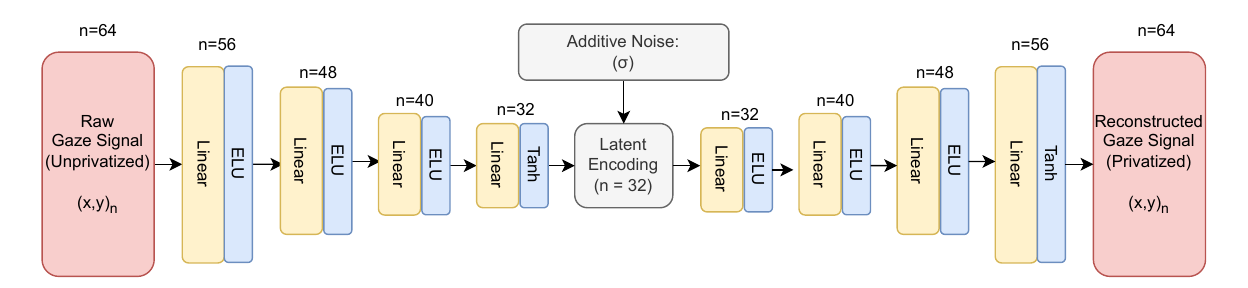}
    \caption{The architecture for the privacy-enhancing autoencoder mechanism.}
    \label{fig:ae-architecture}
\end{figure*}

Figure~\ref{fig:ae-architecture} shows an overview of the model architecture, which contains three major components: the encoder, the additive noise parameter, and the decoder.

The encoder consists of three linear layers that map the gaze signal input from a size of 64 samples to 32 features, followed by an Exponential Linear Unit (ELU) activation function.
Another linear layer followed by a Tanh activation function constrains the values of the latent representation to [-1,1] to encourage consistency in the latent space.
Because the gaze signal is compressed during the encoding process, only the most important components that retain the overall structure of reconstructed signal are retained, which effectively eliminates user-specific features.

To further obfuscate these user-specific features in the latent representation, we include a tunable noise parameter $\sigma$ that adds Gaussian noise to the latent encoding.
This $\sigma$ parameter injects Gaussian noise sampled from a distribution of $(0,\sigma)$ directly into the latent encoding, which randomly perturbs the gaze signal's latent representation and makes it difficult to consistently re-construct the same input gaze sample given a noised encoding.
Increasing the value of $\sigma$ increases the scale of noise that is added, which affords more privacy protection at the cost of a noisier gaze reconstruction. 
This additive noise parameter allows us to designate different levels of privacy protection based on the user's utility needs, which is further explored in Section~\ref{sec:put}.

The decoder reconstructs the original gaze signal from the latent representation by sampling from the encoding, then feeding the encoded representation through three linear layers followed by an ELU activation function. 
A fourth linear layer reconstructs the signal back up to its original input size; we follow this with a Tanh activation function to constrain the output values of the model to a range of [-1,1] to facilitate physiologically plausible reconstructions of the gaze signal.
The gaze signal can then be reconstructed from this output by taking its inverse sine and converting it back into degrees of visual angle.

Because of the bottleneck structure of the autoencoder architecture, we expect some error in the form of lossy reconstruction, as the details of any input is lost during the encoding process.
We also expect additional reconstruction error as a result of perturbing the encodings with the privacy-enhancing noise parameter. 
Ideally, the reconstructed gaze signals retain the overall structure of the original gaze signal, but is absent of personally identifying information.
To achieve this, we designed the training paradigm to encourage the decoder to effectively reconstruct gaze signals from both unperturbed and noised latent encodings.
To train the decoder to effectively reconstruct gaze signals from the noisy privatized latent representations, we include noisy latent representations at training time.
Early in the training process, we allow the autoencoder to learn to reconstruct gaze signals from unperturbed latent embeddings by disabling addition of noise via $\sigma$.
Later in the training process, we gradually inject increasing amounts of noise ranging from 0.0 to 0.1 units into the latent representations before providing them to the decoder.

To encourage the model's robustness to the noise that is injected into the latent representation, designed a loss function that penalizes errors in reconstruction related to discrepancies in gaze position ($\ell_{MSE}$ and $\ell_{DTW}$) and on the overall robustness of the model to perturbations ($\ell_{FGSM}$).
\begin{equation}
    L = \ell_{FGSM} + \ell_{MSE} + \ell_{DTW},
\end{equation}
where $\ell_{FGSM}$ is an adversarial perturbation of the using the fast gradient sign method from Goodfellow et al.~\cite{fgsm}, $\ell_{MSE}$ is the mean square error (MSE) between the original and reconstructed gaze signal, and $\ell_{DTW}$ is soft dynamic time warping discrepancy between the original and reconstructed gaze signal~\cite{dtw}.

The model was trained for 50 epochs with learning rate of .0003 and a batch size of 256 sequences.
All development, training, and testing took place on a Lambda Labs server equipped with dual NVIDIA RTX A5000 GPUs and an AMD Ryzen Threadripper Pro 32-Core CPU @ 2.2GHz with 64 GB RAM.

\subsection{Privacy Assessment}
\label{sec:ekyt}
In this problem setting, a privacy mechanism is evaluated by its ability to remove personally identifying features from the gaze signal.
An effective privacy mechanism would perturb a gaze signal in a way that makes it more difficult to re-identify a subject, thus degrading the performance of a biometric authentication method. 
We use the EyeKnowYouToo (EKYT) biometric authentication model~\cite{Lohr2022} to evaluate our autoencoder mechanism in this context.
As the state-of-the-art biometric authentication model for eye movement, EKYT is commonly employed in the evaluation of privacy-enhancing technologies for gaze data~\cite{wilson2024privacypreserving, Prass2023_synthesis}.

We first train EKYT using the same training-testing split as our autoencoder mechanism (Section~\ref{sec:data}). 
Namely, we train EKYT on the 217 subjects present in the training and testing sets.
Because EKYT uses four-fold validation, we did not have to designate a held-out validation set---these subjects were simply included in the training set. 

After re-training the model, biometric performance was then evaluated using the same set of 105 subjects used to evaluate the autoencoder mechanism.
We use data from each test subject's Session 1 RAN task to produce their enrollment templates, and use their Session 2 RAN data for authentication.
Biometric performance is measured in terms of Equal Error Rate (EER) and Rank-1 identification rate (IR), to represent performance during both biometric verification and identification.
Ideally, an effective privacy mechanism degrades EKYT's biometric performance to near-chance-level authentication performance.
For biometric verification, this corresponds to 50\% EER.
For biometric identification, chance-level performance is $1/105 \approx$ 0.95\% Rank-1 IR. 
The privacy afforded by a given mechanism is higher as biometric performance produced by EKYT approaches chance-level.

\subsection{Utility Assessment}
To assess the utility of a transformed gaze signal, it is important to evaluate how well the signal retains its usefulness for typical downstream applications.
We define utility as the extent to which the privatized signal remains similar to the original and continues to support tasks commonly performed using gaze data in consumer devices.

First, we describe utility degradation of the gaze signal using the mean square error (MSE) between the original gaze signal and its privatized counterpart.
MSE provides a straightforward spatial comparison that can be calculated without any auxiliary information. 
A low MSE indicates that the gaze signal remains close to the original, suggesting that essential spatial characteristics of eye movements have been preserved. 
This is especially important because significant distortions can reduce compatibility with existing models or tools that rely on unmodified gaze data---a privatized gaze signal that retains it resemblance to the original data can be readily integrated into third party tasks that were likely developed on ``normal'' gaze signals, which makes the privacy mechanism more straightforward to adopt.

Second, we observe a task-specific indicator of utility through a benign application that gaze data is often used for in consumer devices: gaze prediction~\cite{Patney2016}.
We follow the approach of Aziz et al.~\cite{aziz2023practical}, employing a recurrent neural network (LSTM) to predict future gaze positions 60 milliseconds ahead, based on recent gaze history.
The gaze prediction error serves as our measure of utility, and is calculated as the Euclidean distance between the predicted gaze prediction produced by the model and the actual position of the gaze at the forecasted time. 
Although it is expected to observe higher gaze prediction errors with increasing levels of privacy, an ideal privacy mechanism would produce limited degradation and retain its overall usability for this application as the value of~$\sigma$ increases.
Note that we utilize the same subject-disjoint train-test split described in Section~\ref{sec:data} to train this model, as well.

\subsection{Privacy-Utility Trade-off}
\label{sec:put}
To evaluate the privacy-utility trade-off of our autoencoder mechanism, we produce privatized gaze data across three levels of privacy protection corresponding to the noise scale $\sigma$ that was injected into the latent representation of the encoded gaze prior to reconstruction.
Privatized gaze data is produced for the full set of 322 GazeBase subjects to facilitate the training and evaluation of our privacy and utility models, and only the subjects in the testing set were used for evaluation.
For each set of privatized data, we train and test a version of EKYT to gauge biometric performance. 
We also calculate the MSE between each data set and the corresponding unperturbed counterpart (``raw'' data), and train and evaluate a gaze prediction model.

We designate a low-privacy setting (``AE-None''), a medium-privacy setting (``AE-0.1''), and a high-privacy setting (``AE-0.2'') based on experimentation--these settings respectively correspond to not adding any noise to the latent representation, adding noise with a scale of $\sigma=0.1$, and adding noise with a scale of $\sigma=0.2$ respectively.
Although $\sigma$ can theoretically take any value, we selected the values above based on the noise levels used during autoencoder training. Specifically, AE-None and AE-0.1 correspond to noise scales seen during training, while AE-0.2 represents a significantly higher noise level than the model was trained on. This setup allows us to evaluate the autoencoder’s denoising capability (and thus its ability to preserve utility) when faced with noise levels beyond its training range.
There likely exist several other values of $\sigma$ that are useful for privacy enhancement---these values would typically be chosen based on whether their corresponding privacy-utility trade-offs satisfy the utility needs of a given application.  

\section{Results}
Tables~\ref{tab:put-ran} summarizes the privacy-utility trade-off measured for the RAN task.
From these summary values, the autoencoder mechanism effectively decreases biometric authentication performance close to chance-level while maintaining the overall utility of the data.
Firstly, the autoencoder mechanism provides significant privacy enhancement by obscuring the personally identifying information that EKYT needs to successfully re-identify users across play sessions.
This is evidenced by the degradation of EER from 9.4\% to 38.9\% (toward chance-level performance of 50\%), and the degradation of Rank-1 IR from 81.9\% to 10.5\% (toward chance-level performance of 0.95\%). 
Each level of privacy protection also produces increasingly degraded biometric authentication rates, which indicates that injecting noise into the autoencoder's latent representations is an effective way to impart user privacy.

We also observe relatively low levels of degradation in the utility of the data relative to the amount of privacy enhancement across privacy levels, which indicates the autoencoder's ability to robustly reconstruct gaze signals from these latent representations, even in the presence of the privacy-enhancing noise injection.
While the MSE relative to the original signals across privacy levels increase gradually, the gaze prediction error of the privatized data actually decreases on average; the latter effect is an unusual observation that explore in the following section.

\begin{table}[]
    \centering
    \caption{Summary privacy-utility trade-off across privacy levels. Arrows indicate the direction of better performance.}
    \label{tab:put-ran}
    \begin{tabular}{|c|r|r|r|r|}
    \hline
     & \multicolumn{2}{c|}{Privacy Measure} & \multicolumn{2}{c|}{Utility Measure (\degree)} \\
    \makecell{Privacy \\ Level} & \makecell{EER \\ (\%, $\downarrow$)} & \makecell{IR \\ (\%, $\uparrow$)} & MSE & \makecell{Prediction \\ Error} \\
    \hline
    Raw & 9.4 & 81.9 & 0.00 & 1.82 \\
    AE-None & 23.8 & 34.3 & 0.09 & 1.17 \\
    AE-0.1 & 30.8 & 14.3  & 0.17  & 1.33 \\
    AE-0.2 & 38.9 & 10.5 & 0.30  & 1.41 \\
    \hline
    \end{tabular}
\end{table}

\subsection{Privacy Enhancement}

\begin{figure*}
    \centering
    \begin{subfigure}[b]{0.375\linewidth}
        \includegraphics[width=\linewidth]{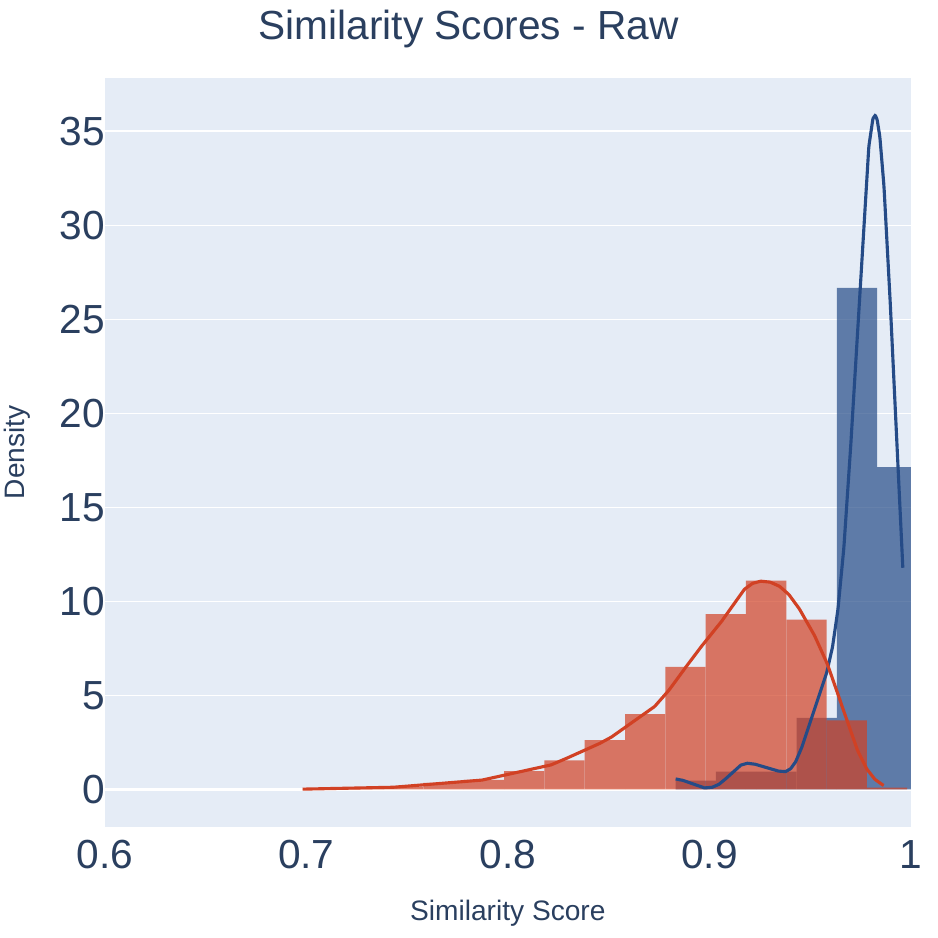}
        \caption{Raw data.}
        \label{fig:raw-dist}
    \end{subfigure}
    \begin{subfigure}[b]{0.375\linewidth}
        \includegraphics[width=\linewidth]{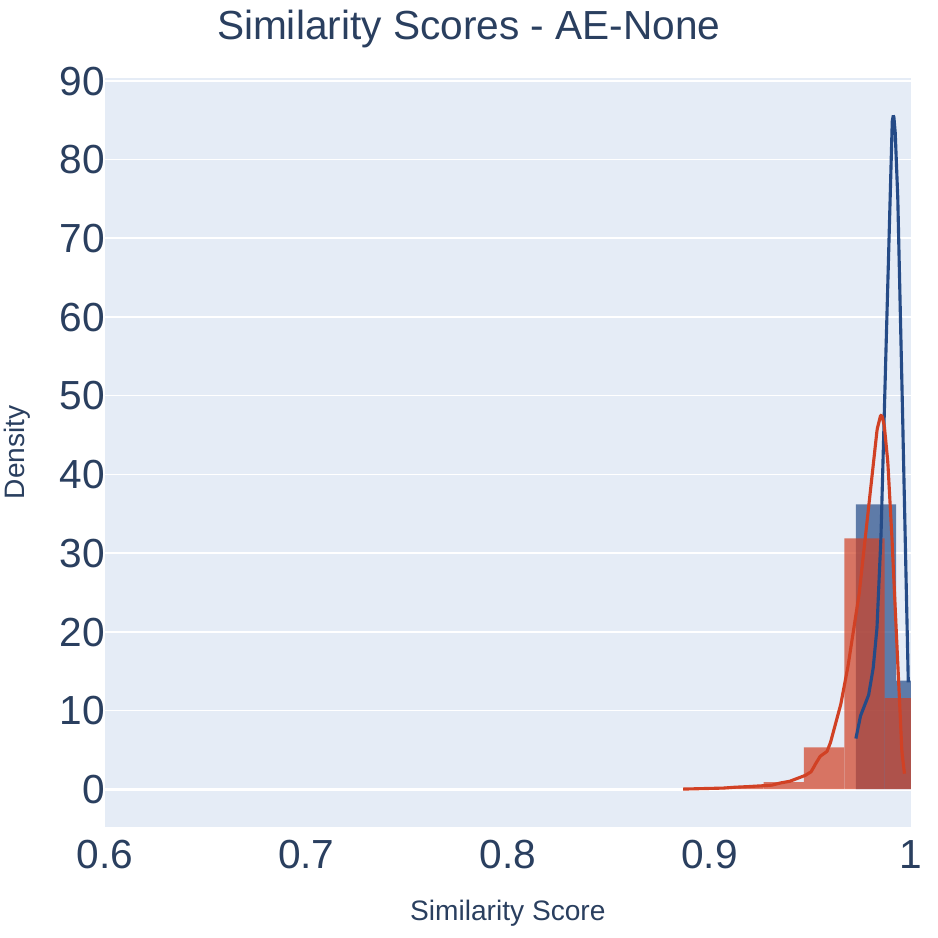}
        \caption{AE-None.}
    \label{fig:none-dist}
    \end{subfigure}
    \\
    \begin{subfigure}[b]{0.375\linewidth}
        \includegraphics[width=\linewidth]{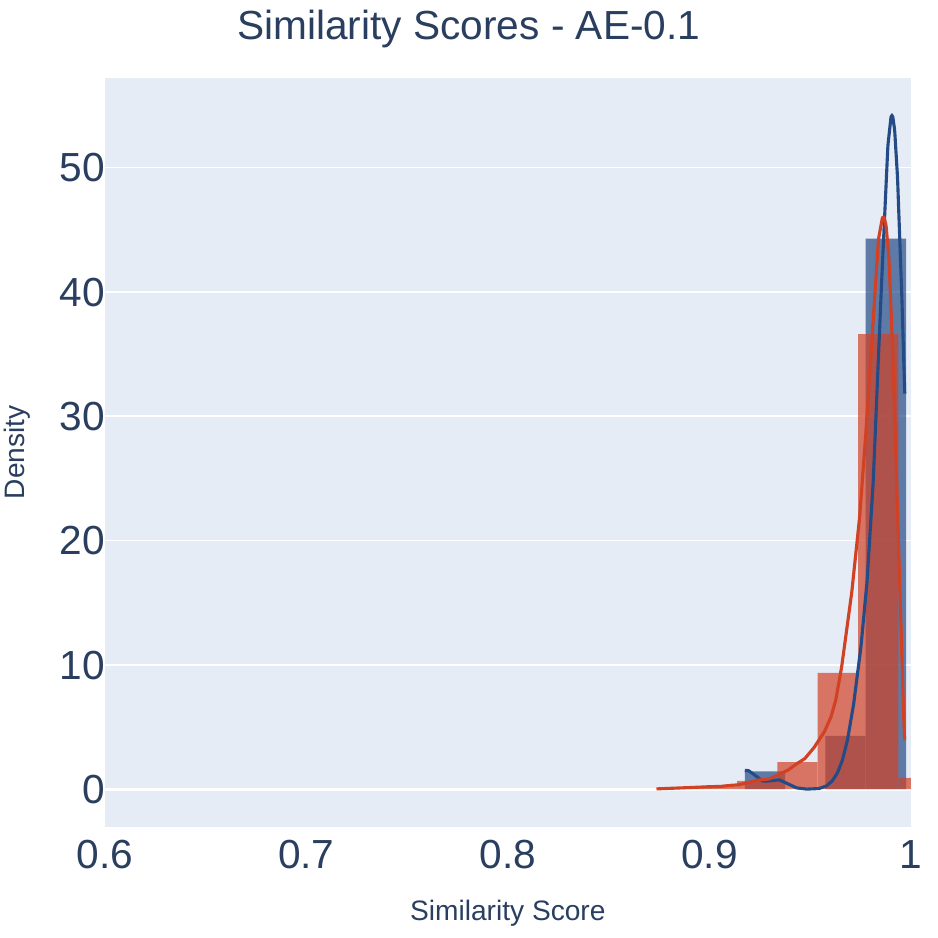}
        \caption{AE-0.1.}
    \label{fig:0.1-dist}
    \end{subfigure}   
    \begin{subfigure}[b]{0.375\linewidth}
        \includegraphics[width=\linewidth]{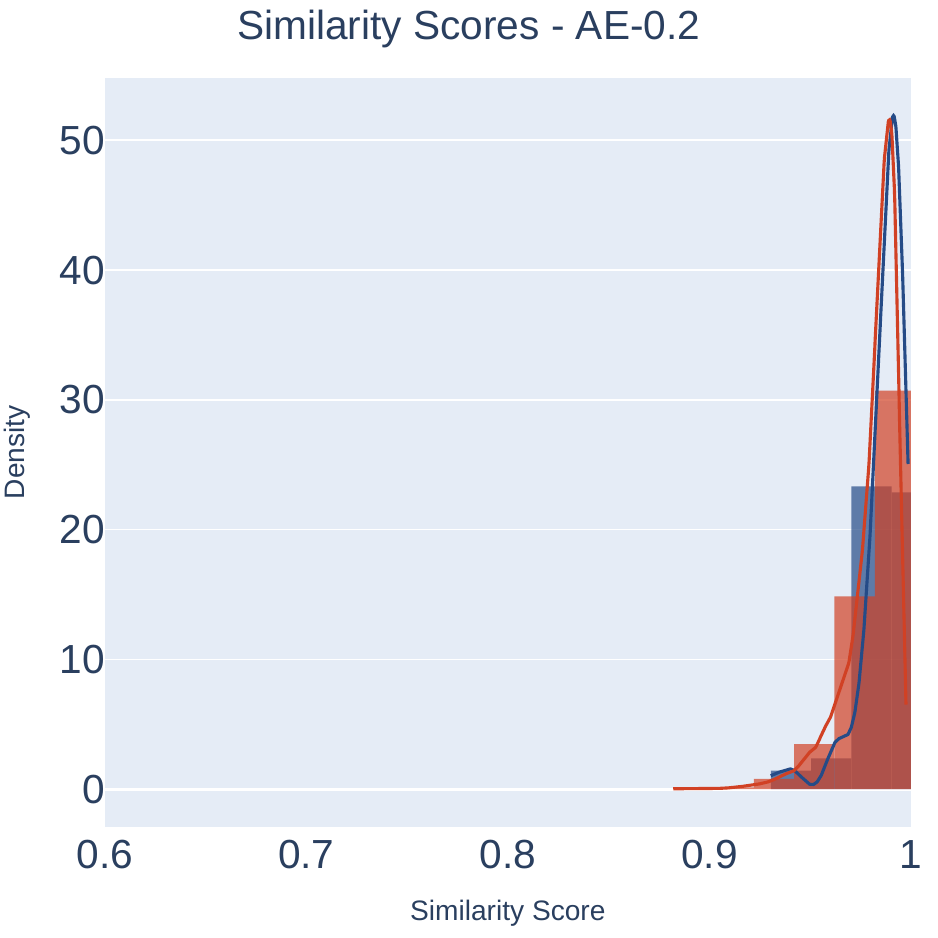}
        \caption{AE-0.2.}
    \label{fig:0.2-dist}
    \end{subfigure}    
    \caption{Distribution of similarity scores across privacy mechanisms. Note that larger separation between the genuine (blue) and imposter (red) distributions is associated with better biometric performance.}
    \label{fig:ch4-similarity-dists}
\end{figure*}

Although the EER and Rank-1 IR metrics reported in Table~\ref{tab:put-ran} can succinctly summarize biometric performance, they may not convey how a privacy mechanism works to degrade biometric performance.
These singular descriptors can be supported by showing the distribution of similarity scores between genuine and imposter authentication attempts, as shown in Figure~\ref{fig:ch4-similarity-dists}.
The similarity score distribution of an effective biometric authentication model features a large separation between genuine and imposter scores, which demonstrates that the model can effectively separate genuine authentication attempts from imposters. 
The similarity score distribution produced with unperturbed data (``Raw'', Figure~\ref{fig:raw-dist}) has an example of such a distribution.

Applying a privacy mechanism to degrade biometric performance changes the similarity distributions so that the imposter scores and the genuine scores have more overlap, making it difficult for the model to distinguish between genuine and imposter authentication attempts.
The degree to which these distributions overlap dictate the model's biometric performance, and are directly related to the EER and Rank-1 IR scores presented in Table~\ref{tab:put-ran}.
As the privacy level increases, we observe more overlap between the genuine and imposter distributions, which indicates that EKYT has greater difficulty successfully identifying whether a gaze sample matches a given enrollment template. 
The emergence of this overlapping effect is best observed in the transition between the Raw data and AE-None, with AE-0.1 and AE-0.2 increasing the degree of overlap to marginally decrease biometric authentication performance.

\begin{figure}
    \centering
    \includegraphics[width=0.45\linewidth]{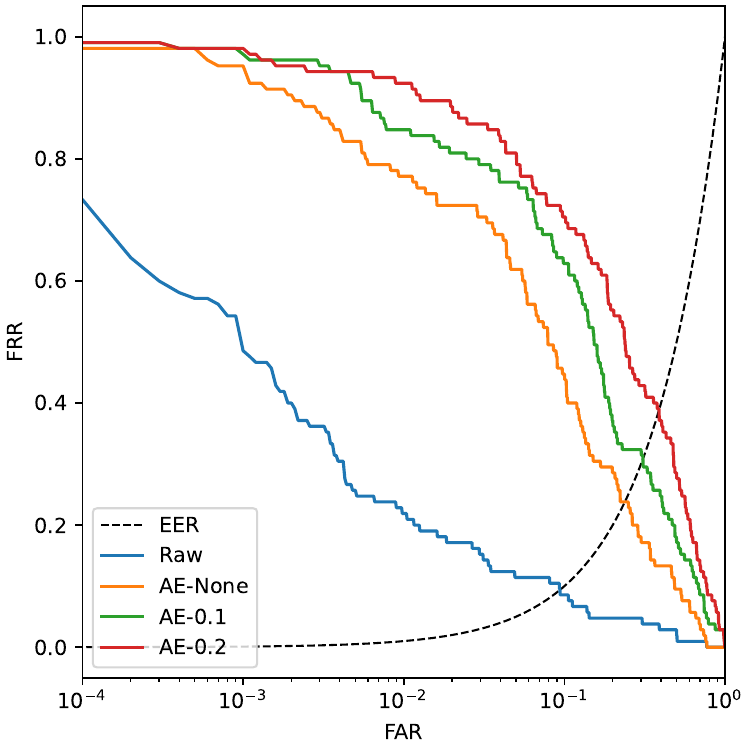}
    \caption{ROC curves across each privacy configuration.}
    \label{fig:roc}
\end{figure}

Figure~\ref{fig:roc} shows the ROC curve for each privacy consideration when evaluating biometric verification rates, and further illustrates that the greatest privacy enhancement comes from the reconstruction loss inherent to the autoencoder, and that adding noise to the encodings provide additional marginal privacy enhancement.

\subsection{Utility Degradation}

\begin{figure}
    \centering
    \includegraphics[width=0.65\linewidth, trim={0 0 3.3cm 0},clip]{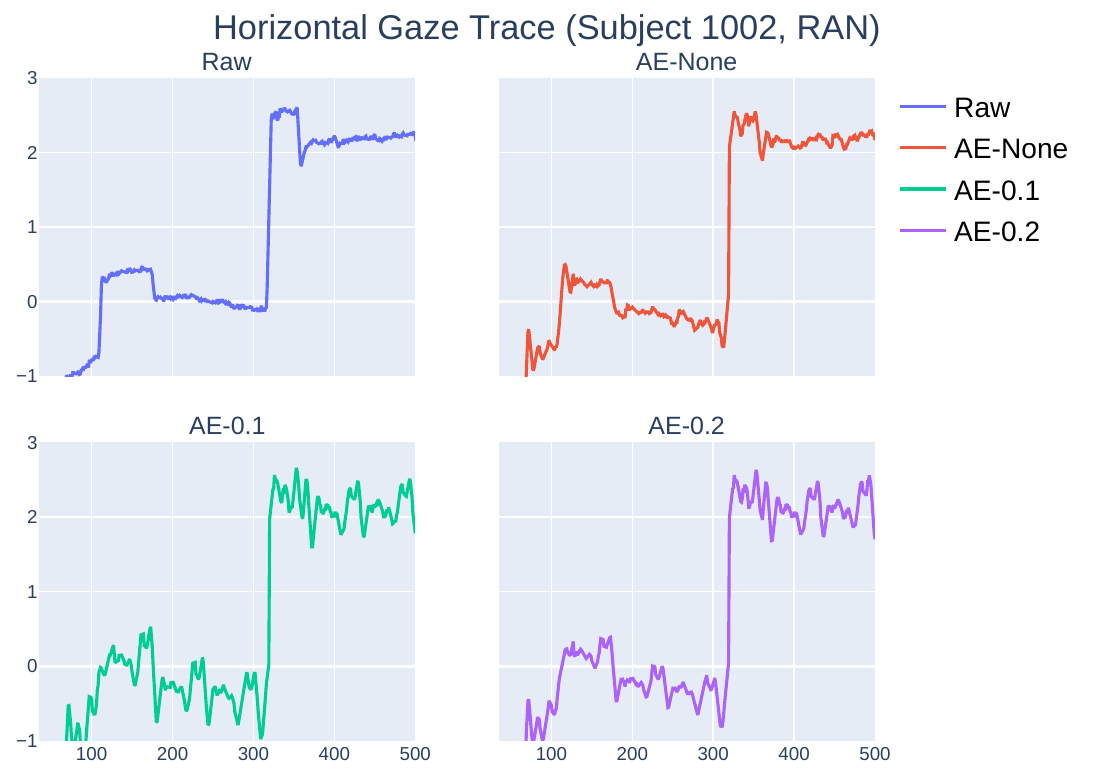}
    \caption{How each privacy level distorts the same gaze signal.}
    \label{fig:signal-viz}
\end{figure}

An intuitive way to gauge the effect of privatization on a gaze signal is to simply visualize it. 
Figure~\ref{fig:signal-viz} shows an example of a single gaze trace after it has been processed at each level of privacy protection.
The raw gaze data (blue) illustrates a clear sequence of four fixations (intervals of relative eye stability) and four saccades (rapid transitory movements between fixation points) captured without additional processing.
In the privatized gaze signals, the overall temporal and spatial structure of the original trace is generally preserved.
However, finer-grained features, such as brief saccades and small fixations, tend to become less distinguishable as the level of privacy increases.
At the highest privacy setting (``AE-0.2''), the perturbation distorts gaze positions by a scale of approximately one degree of visual angle.
This level of distortion is well-within the range of typical measurement error in many commercial eye tracking systems, and can considered acceptable for a variety of practical applications.

\begin{figure}
    \centering
    \includegraphics[width=0.75\linewidth]{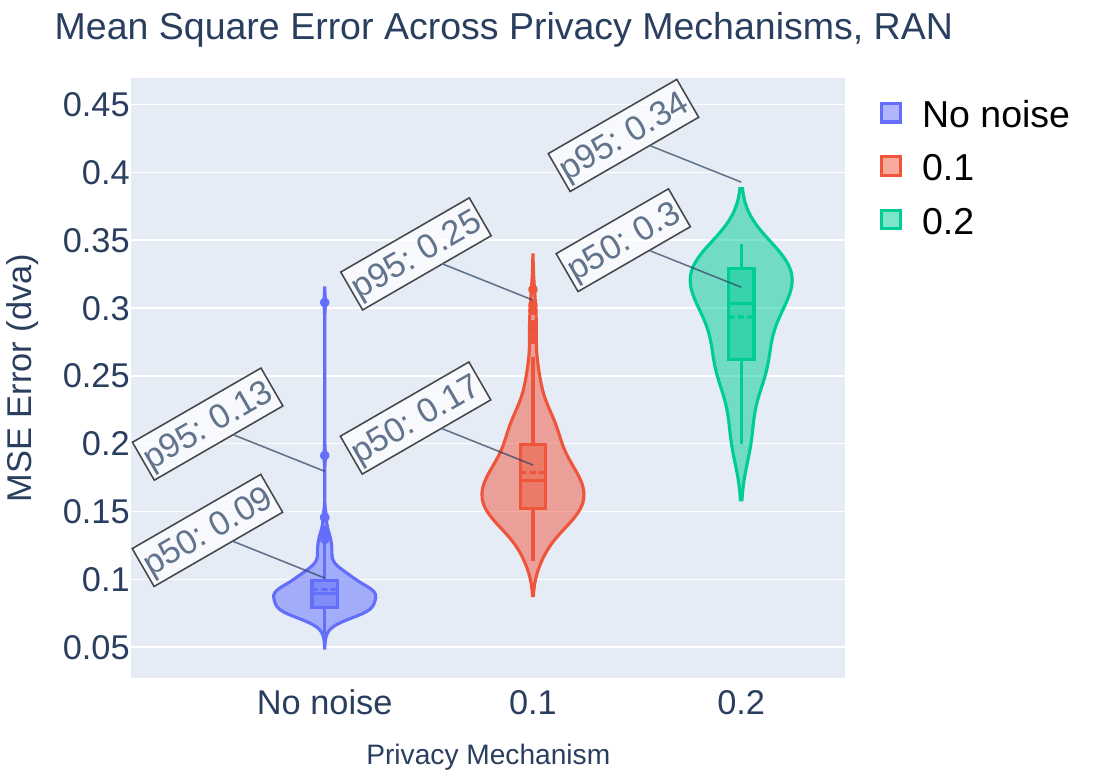}
    \caption{The distribution of mean squared error between the raw data and privatized data across all privacy levels.}
    \label{fig:mse}
\end{figure}

Figure~\ref{fig:mse} shows the distribution of MSE errors across all gaze signals at each privacy level relative to the original, raw data.
The reconstructed gaze signals tend to feature an increase in MSE that roughly corresponds to the value of the noise parameter $\sigma$ that was applied to the latent encodings---AE-0.1, for example, tends to feature an average increase of MSE values of 0.1 relative to AE-None.
The spread of these distributions show that, on average, the privatized gaze signals differ significantly on a sample-to-sample basis across privacy levels. 
Interestingly, the autoencoder tends to increase the range of the distribution of MSE values, as well as eliminate the outliers who might naturally exhibit high levels of noise in their original gaze signals.
This effect may contribute to the obfuscation of personal information, as these subjects would be harder to distinguish based on the levels of noise in their gaze signals alone at higher privacy levels. 

\begin{figure}
    \centering
    \includegraphics[width=0.65\linewidth]{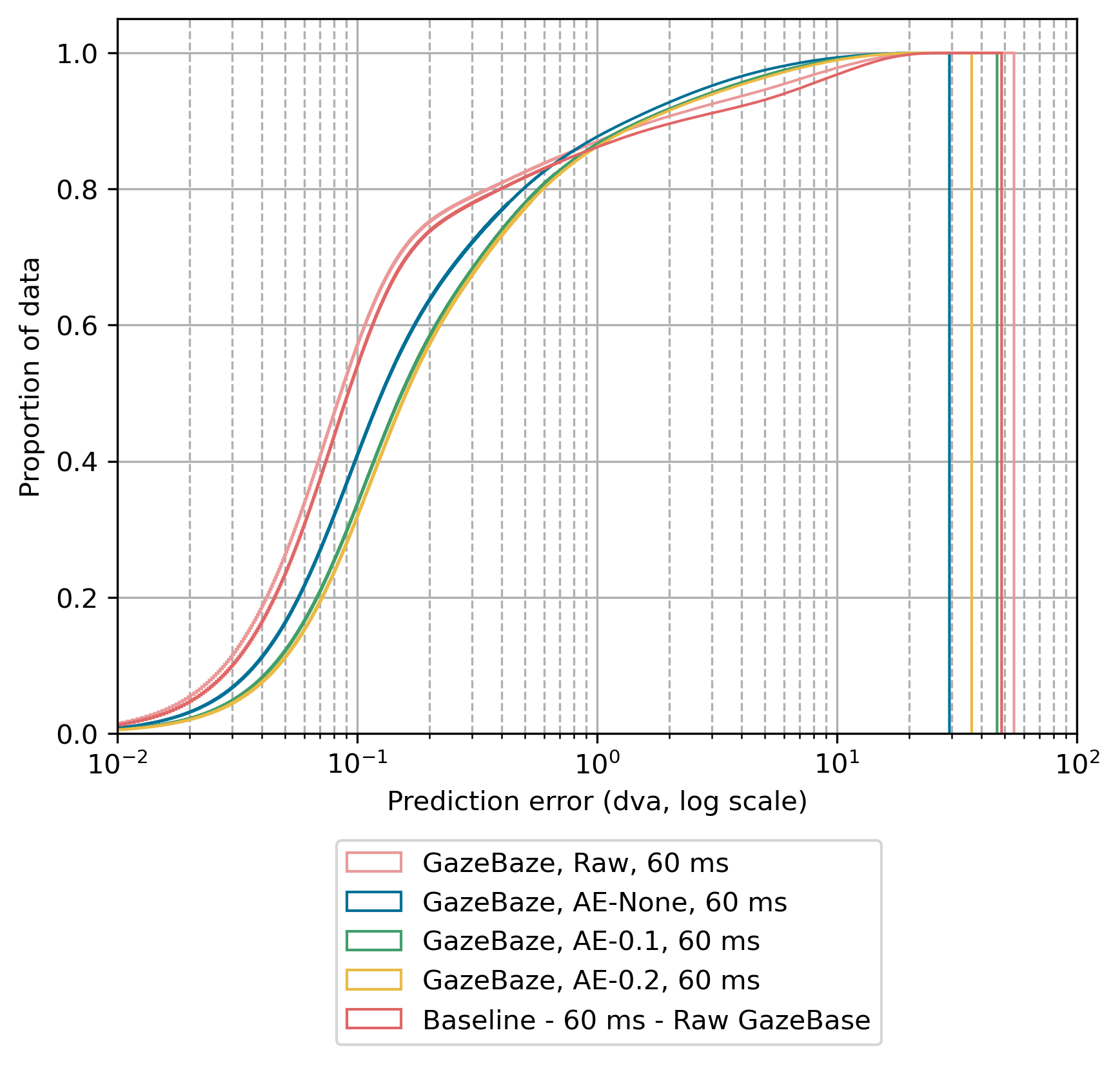}
    \caption{Gaze prediction error across all privacy mechanisms.}
    \label{fig:gazepred-cdf}
\end{figure}

Overall, we do not observe a significant degradation in utility during gaze prediction. 
Gaze prediction performance was evaluated using a cumulative distribution function (CDF) of prediction errors across the entire test set, as shown in Figure~\ref{fig:gazepred-cdf}. 
This CDF captures the proportion of gaze samples (vertical axis) that fall below a given prediction error threshold (horizontal axis), providing a holistic view of prediction accuracy. 
Gaze prediction results are shown for five testing configurations: the raw data, data for each level of privacy, and a baseline condition in which no predictive modeling is applied to serve as a reference point for the quality of prediction.
As the proportion of data increases along the vertical axis (in this case, this measures the proportion of gaze samples), the cumulative prediction error for those samples is tracked along the horizontal axis.
The best-performing gaze prediction featuring the lowest prediction error trend toward the top-left of the graph.

Overall, the transformed gaze signals retain strong utility, with only modest increases in prediction error compared to raw data.
The degradation in utility is gradual across each test condition, with prediction errors increasing by less than half a degree of visual angle on average.
As shown in Figure~\ref{fig:gazepred-cdf}, the prediction error for privatized data is slightly higher for a majority of the data, especially for particularly for the roughly 80\% of gaze samples corresponding to fixations~\cite{aziz2023practical}.
However, the privatized data exhibits better performance on data that is more difficult to predict.
Specifically, during saccadic movements, the models trained on privatized signals yield lower maximum prediction error than those trained on raw data. 
This suggests that while fixational noise increases error in the easier segments of the signal, the predictive model is learning to generalize better to higher-variance behaviors, such as saccades.
This likely contributes to the improved ``average'' gaze prediction errors that we observe in Table~\ref{tab:put-ran}~relative to the raw data.

Gaze prediction performance is likely tied to the amount of noise observed in the privatized gaze signals, as shown in Figure~\ref{fig:signal-viz}.
The raw signal features highly stable fixations with minimal jitter, making them easy to predict by assuming zero displacement. 
In contrast, the privatized signals introduce greater variability during fixations, reducing the effectiveness of these simple predictions and increasing the average error. 
Given the high proportion of fixational data in the test set, this contributes to the slight overall degradation in performance. 
However, the ability of the model to better predict complex dynamics such as saccades indicates that the utility of the privatized signals is still suitable for gaze prediction in a practical setting. 
Note that, as seen in Figure~\ref{fig:gazepred-cdf}, the lowest privacy setting (AE-None) produces the best gaze prediction results of all of the privatized signals, but the difference in error between the medium and high privacy levels (AE-0.1 and AE-0.2) are minimal across the entire proportion of data.

\section{Discussion}
\label{sec:discussion}

Overall, the autoencoder mechanism offers a favorable privacy-utility trade-off by reducing the rates of biometric identification while maintaining the utility of gaze data for benign downstream tasks. 
Balancing privacy and utility is an important indication of success; other methods that have been proposed for eye tracking privacy tend to produce privatized gaze signals that feature significant loss in data utility or produce gaze signals that deviate from physiologically plausible eye movement patterns~\cite{Hu2022_otus,davidjohn2021,wilson2024privacypreserving}.
One significant advantage of our proposed mechanism is that it was explicitly designed to preserve the structure of the eye movement signal by significantly penalizing deviations from physiologically plausible outputs.
In contrast, prior approaches that utilize additive noise mechanisms do not include safeguards for maintaining this structure, especially if noise is sampled separately for the horizontal and vertical components of the gaze~\cite{davidjohn2021}.
This can lead to synthetic gaze signals that violate known physiological constraints, thereby impairing their usability. 
Our autoencoder addresses this shortcoming by learning to reconstruct gaze signals that are structurally faithful to real-world eye movement patterns.

Interestingly, the greatest degree of privacy protection comes from the reconstruction loss of the autoencoder itself, possibly because the majority of personally identifying information in the gaze signal is already eliminated through the autoencoder's bottleneck compression. 
However, adding noise to the encodings is still worthwhile because it provides controllable privacy enhancement that would not be available otherwise.
Relying on the outcome of a stochastic training method (in this case, reconstruction loss) to enhance privacy introduces too much uncertainty to make the privacy mechanism useful in practical settings, as it can be difficult to evaluate and tune reliably.
Adding known noise to the encodings can enable a more consistent degree of privacy enhancement, which complements the privacy-enhancing effect of a standalone autoencoder.

We did not observe chance-level privacy enhancement with this mechanism, which indicates that some personally identifying information is still present in the privatized gaze signals. 
One possible explanation for this is that the autoencoder mechanism only perturbs spatial domain of the gaze signal, which only privatizes personally identifying information that is encoded in gaze positions.
Gaze data also contains subject-specific information in the temporal dimension of a user's gaze dynamics; for example, characteristics like reaction time~\cite{lee2019differences} and person-specific changes due to cognitive load~\cite{krejtz2018eye} might convey personally identifying information that this mechanism does not address.
This is a key limitation of our proposed approach, as it has direct implications on how much privacy can be afforded to users.

\section{Future Work}
\label{sec:future}
To further advance privacy protection for gaze data, future work might explore privacy enhancing mechanisms that address both the spatial and temporal aspects of personally identifying information in gaze data.
While some work in this direction exists for event-level privacy for real-time gaze streaming~\cite{li2021}, the development of user-level privacy protection in this setting remains an open area of research.
Incorporating a temporal perturbation also has implications for the usability of the privatized gaze signal (especially for time-sensitive applications), and would necessitate the development of utility measures that can adequately assess this complexity (for example, one may consider replacing the time-sensitive MSE measure of utility with a time-invariant counterpart, like dynamic time warping~\cite{dtw}).

\section{Conclusion}
We introduced an autoencoder-based frame work for enhancing user privacy, specifically formulated to reduce the amount of personally identifying information in gaze data.
Through a combination of structural learning and controlled perturbation in the latent space, our method significantly reduces biometric identification performance while maintaining utility for a benign downstream applications that do not require personal information, producing a favorable privacy-utility trade-off by producing realistic reconstructions of real gaze patterns.

Our results offer a promising step toward practical, real-world deployment of privacy-preserving eye tracking in extended reality and other human-centered computing environments.
Future work will explore extensions to temporally aware privacy transformations and formal guarantees for privacy in sequential gaze data.

\section{Acknowledgements}
This work is supported by the National Science Foundation Graduate Research Fellowship (DGE-1840989).

\bibliographystyle{plain}
\bibliography{template}

\end{document}